\documentclass[11pt]{article}

\usepackage[utf8]{inputenc} 
\usepackage{graphicx}       
\usepackage{amsmath}        
\usepackage{amssymb}        
\usepackage[a4paper, margin=1in]{geometry} 
\usepackage[auth-lg]{authblk} 
\usepackage{hyperref}       
\hypersetup{
    colorlinks=true,
    linkcolor=blue,
    filecolor=magenta,      
    urlcolor=cyan,
}
\usepackage{lipsum} 
\usepackage{booktabs} 

\title{\textbf{AURA: A Hybrid Spatiotemporal-Chromatic Framework for Robust, Real-Time Detection of Industrial Smoke Emissions}}
\author[1,3]{Mikhail Bychkov}
\author[1,3]{Matey Yordanov}
\author[2,3]{Andrei Kuchma}

\affil[1]{Department of Integrative Systems and Design, Hong Kong University of Science and Technology}
\affil[2]{Institute of Applied Computer Science, ITMO University}
\affil[3]{AI Innovation Hub, ANTEI Limited}
\affil[ ]{\textit{Email: \href{mailto:msyordanov@connect.ust.hk}{msyordanov@connect.ust.hk}, \href{mailto:akuchma@anteihk.com}{akuchma@anteihk.com}, \href{mailto:mbychkov@anteihk.com}{mbychkov@anteihk.com}}}

\date{\today} 

\begin{document}

\maketitle

\begin{abstract}
This paper introduces AURA (A Hybrid Spatiotemporal-Chromatic Framework), a novel system for robust, real-time detection and classification of industrial smoke emissions. Conventional monitoring systems and standard deep learning models struggle with the challenges of industrial environments, including distant, low-resolution targets, significant camera instability, and the difficulty of distinguishing smoke from atmospheric phenomena like steam and fog. AURA overcomes these limitations through a dual-engine architecture. A spatiotemporal engine isolates true plume motion using an innovative noise cancellation technique that decouples it from camera shake, while a parallel chromatic engine analyzes spectral deviations in the perceptually uniform CIELAB color space. This hybrid approach not only enhances detection accuracy but also enables actionable classification of smoke by type (e.g., black, white), addressing a critical gap in existing systems. Validated on an extensive 1,100-hour dataset of real-world industrial video, AURA achieved a 98.80\% F1-score, significantly outperforming conventional models and demonstrating remarkable resilience under severe environmental conditions. The framework provides a scalable and effective solution for enhancing environmental compliance, operational safety, and industrial process control.
\end{abstract}

\section{Introduction}
\subsection{The Critical Imperative of Industrial Smoke and Emissions Monitoring}
Industrial smoke and emissions pose significant and multifaceted threats to human health, environmental sustainability, and operational safety. These emissions are a primary source of air pollution and are directly linked to adverse health impacts globally, contributing to a range of respiratory and cardiovascular diseases \cite{kelly2015, imran2023}. Beyond direct health concerns, industrial emissions, including particulate matter and byproducts of coal fires, contribute to broader environmental degradation and challenge the sustainability of healthcare infrastructure \cite{imran2023}. The resulting pollution from these industrial activities necessitates robust environmental supervision and effective policy to mitigate its societal and economic costs \cite{wen2022}. \newline

Regulatory bodies have established stringent frameworks to control emissions \cite{martinelli1984}, yet a significant gap persists between regulatory intent and effective, real-time enforcement. The advancement of Artificial Intelligence (AI) presents a significant opportunity to enhance environmental monitoring, offering novel tools for tracking and analyzing industrial emissions with greater precision \cite{olawade2024, chauhan2025}. Computer vision (CV) techniques, in particular, can empower both regulators and citizen scientists by providing objective, data-driven visual evidence from continuous video streams. This confluence of regulatory need and technological advancement indicates that intelligent monitoring systems are critical tools for closing the enforcement gap and fostering proactive air quality management.

\subsection{Limitations of Conventional and Standard CV Approaches}
Traditional emissions monitoring, relying on point sensors and CEMS, is hampered by inherent limitations, including detection delays and restricted spatial coverage \cite{gragnaniello2024, li2020}. While modern CV offers a promising alternative, the direct application of standard deep learning architectures has proven insufficient for the specific challenges of real-world industrial monitoring.\newline

Initial investigations for this work revealed that common CV models fail when applied to distant emission sources. For instance, standard Convolutional Neural Network (CNN) models, even when trained on thousands of labeled frames, perform poorly on remote objects due to the extremely low pixel count available for analysis. Furthermore, models designed to interpret depth or distance have shown an inability to reliably distinguish faint or intermittent emissions from atmospheric phenomena like clouds or fog, leading to unacceptable false alarm rates. Even more complex hybrid models, such as those combining LSTMs with CNNs to analyze motion, struggle to isolate smoke against dynamic and cluttered industrial backgrounds. These tested approaches were ultimately rejected, demonstrating that off-the-shelf solutions are not robust enough for this specific problem domain. This highlights a critical gap: the need for a specialized framework that can operate effectively with low-resolution targets and complex, dynamic scenery.

\subsection{The Transformative Potential of Specialized AI in Emissions Detection}
The rapid advancements in AI and CV offer a transformative paradigm for environmental monitoring, provided that solutions are tailored to specific operational challenges \cite{yuan2020, olawade2024}. Unlike generic models, specialized image-based systems can be trained to analyze subtle visual cues within video streams—such as unique motion patterns and chromatic signatures—to enable significantly earlier and more reliable detection \cite{hu2025, yang2024}. By moving beyond simple object recognition and incorporating spatiotemporal context, AI-driven systems can achieve the granularity needed for effective industrial oversight \cite{chauhan2025, r2025}. This evolution from reactive alerts to proactive, intelligent analysis represents a fundamental shift towards preventative environmental protection, powered by algorithms designed for the nuances of their target environment.

\subsection{Enduring Challenges in Advanced Industrial Smoke Detection}
The development of a truly robust industrial smoke detection system is confronted by several persistent challenges, which were confirmed during the preliminary phases of this research:
\begin{itemize}
    \item \textbf{Target Scale and Environmental Interference:} Emission sources are often located at a great distance from the camera, resulting in a target with a very small pixel footprint. This low resolution makes feature extraction extremely difficult. Furthermore, real-world environmental factors like camera shake, wind, and variable lighting severely degrade model performance \cite{hu2025, gragnaniello2024}.
    \item \textbf{Distinguishing Smoke from Non-Smoke Phenomena:} A primary challenge is the high visual similarity between industrial emissions and non-threat phenomena like steam, clouds, and fog. This is a frequent source of false positives, particularly for faint emissions that blend with the background \cite{yang2024}.
    \item \textbf{Data Scarcity and Specificity:} While many image datasets exist, there is a critical lack of large-scale, labeled video datasets that capture the dynamic, temporal nature of industrial emissions. The dataset for this project consists of continuous video recordings, processed in 3-second intervals, to specifically address the need for temporal analysis.
    \item \textbf{Computational Requirements for Real-time Operation:} Real-time analysis of multiple high-resolution video streams demands computationally efficient algorithms that can be deployed without prohibitive hardware costs \cite{kong2024}.
    \item \textbf{Lack of Smoke Type Classification:} Critically, the vast majority of existing systems are binary; they detect the presence or absence of smoke but fail to classify its type (e.g., black, white, smokeless). This classification is essential for regulatory compliance and operational process control, as smoke color is a direct indicator of combustion efficiency and environmental impact.
\end{itemize}

\subsection{Motivation for AURA: Towards Robust, Real-Time, and Type-Specific Detection}
The identified failures of standard CV models and the persistent challenges in the field directly motivate the development of the AURA framework. AURA is founded on a novel hybrid spatiotemporal-chromatic approach designed explicitly to overcome these limitations. The core of the system is an ensemble of motion-based and color-based segmentation models that analyze dynamic features within 3-second video intervals. \newline

The algorithm operates by isolating a "box" around a potential emission source and analyzing changes in motion and color relative to the surrounding background. This differential analysis, which incorporates a Mixture of Gaussians method, allows the system to distinguish true emissions from global changes like shifting light or clouds. The spatiotemporal component analyzes the unique, evolving movement patterns of smoke, while the chromatic component provides precise, color-based categorization. This hybrid architecture was chosen after systematically proving the inadequacy of other approaches, with the goal of creating a system that is robust, real-time, and provides the specific, actionable intelligence required for modern industrial monitoring.

\subsection{Contributions of This Study}
This study introduces AURA, a novel framework for industrial emissions detection that makes the following key contributions:

\begin{itemize}
    \item \textbf{A Hybrid Spatiotemporal-Chromatic Architecture:} We propose a unique ensemble of motion and color-based segmentation models that provides robust detection by analyzing dynamic features within short video intervals, a method proven superior to standard CNN or LSTM-based approaches for this task.
    \item \textbf{Actionable Smoke Type Classification:} AURA incorporates a chromatic analysis module that classifies emissions into distinct categories (black smoke, white smoke, smokeless), moving beyond simple presence/absence detection to provide critical data for regulatory compliance and process optimization. 
    \item \textbf{Novel Camera Shake Stabilization:} We introduce a "satellite box" stabilization technique, where a secondary background region is used to calculate and subtract camera shake. This method is specifically designed to maintain metric stability for distant, low-pixel targets where traditional stabilization algorithms fail.
    \item \textbf{Real-World Validation:} The AURA framework is validated on a large-scale dataset composed of continuous, real-world video recordings from an industrial site, demonstrating its effectiveness and robustness in a true operational environment, not just on curated image benchmarks.      
\end{itemize}

\section{Related Works}
\label{sec:related_works}

\subsection{Traditional Smoke and Emissions Monitoring Systems}
Historically, industrial emissions monitoring has relied on conventional technologies such as point sensors (e.g., photoelectric or ionization-based detectors) and Continuous Emission Monitoring Systems (CEMS). Point sensors operate by detecting physical changes at a fixed location, but their application in large, open industrial environments is severely limited by their narrow field of view and the significant time required for smoke to travel to the sensor \cite{li2020}. This inherent latency and the high installation costs required for comprehensive coverage make them unsuitable for early-warning systems in expansive areas \cite{gragnaniello2024}. \newline

CEMS are widely used by environmental agencies to measure the concentration of specific gaseous pollutants and particulate matter. While providing valuable quantitative data, these systems offer no visual context, cannot pinpoint specific emission sources within a complex facility, and are unable to classify smoke based on its visual characteristics \cite{gragnaniello2024}. Both approaches struggle with environmental interference and, critically, fail to provide the visual, qualitative information - such as smoke color, density, and dynamic behavior - that is vital for assessing combustion efficiency and making immediate operational adjustments. This data modality mismatch means that operators often lack the intuitive, real-time feedback required for precise process control, highlighting the need for computer vision as a necessary evolution in monitoring technology.

\subsection{Early Computer Vision Approaches for Smoke Detection}
Early computer vision techniques for smoke detection relied on hand-crafted features to model the physical properties of smoke. These methods often focused on motion, color, and texture analysis. Motion-based approaches utilized techniques like optical flow to identify the characteristic turbulent movement of smoke, while others employed background subtraction, modeling each pixel as a mixture of Gaussians to segment moving regions \cite{lee2009}. Color and texture analysis were also common, leveraging the typical gray or white appearance of smoke and exploiting cues from contours and regional brightness \cite{lee2009}. More sophisticated early methods combined these spatial and temporal characteristics, using classifiers like Support Vector Machines (SVMs) to integrate features such as edge blurring and gradual chromatic changes over time \cite{lee2009}.

While foundational, these methods were often computationally intensive and highly sensitive to environmental conditions like rapid lighting changes \cite{hu2025}. Their reliance on predefined features resulted in a critical ``feature engineering bottleneck,'' as manually designing features robust enough to capture the complex, variable appearance of smoke across diverse industrial conditions proved intractable. This led to poor generalizability and high false alarm rates, particularly in distinguishing smoke from visually similar phenomena like steam or fog, preventing their widespread practical deployment and paving the way for data-driven deep learning approaches.

\subsection{Deep Learning-Based Smoke Detection}

\subsubsection{Overview of Deep Learning Architectures}
The advent of deep learning, particularly Convolutional Neural Networks (CNNs), revolutionized object detection by automating the feature extraction process. Modern smoke detection has heavily leveraged object detection frameworks, which are broadly categorized into two types. Two-stage detectors, such as Faster R-CNN, first generate region proposals and then classify them, offering high localization accuracy but at a greater computational cost \cite{gragnaniello2024}. In contrast, one-stage detectors like the You Only Look Once (YOLO) family and Single Shot MultiBox Detector (SSD) perform direct prediction, achieving a better balance of speed and accuracy for real-time applications \cite{li2020, gragnaniello2024}. Advanced versions like YOLOv8, often enhanced with attention mechanisms, have served as powerful baselines for real-time fire and smoke detection systems \cite{kong2024, hu2025}.

To address the challenges of variable object scales and complex scenes, researchers have integrated multi-scale feature fusion and contextual information into these architectures \cite{hu2025}. More recently, Transformer-based models like RT-DETR have shown promising results by leveraging attention mechanisms to outperform previous baselines in smoke detection tasks \cite{yang2024}. These trends point towards increasingly sophisticated models that can process complex visual data with high efficiency.

\subsubsection{Challenges and Limitations in Deep Learning for Smoke Detection}
Despite significant advancements, deep learning-based smoke detection faces persistent challenges that limit its real-world industrial applicability:
\begin{itemize}
    \item \textbf{Data Dependency and Scarcity:} Deep learning models are ``data-hungry'' and require large, diverse, and high-quality training datasets to achieve robust performance \cite{yuan2020}. The lack of specialized industrial smoke datasets remains a major bottleneck, often leading to overfitting and poor generalization \cite{gragnaniello2024}.
    \item \textbf{False Alarms and Environmental Robustness:} A primary concern is the high rate of false alarms, where models misinterpret non-fire events such as steam, fog, dust, or rapid lighting changes as smoke \cite{yang2024, hu2025}. The highly variable shape and motion of smoke, especially when affected by wind, exacerbate this issue \cite{gragnaniello2024}.
    \item \textbf{Computational Overhead:} Achieving real-time performance with complex models requires substantial computational resources, posing a challenge for deployment on resource-constrained edge devices \cite{kong2024, hu2025}.
    \item \textbf{Scale Variability and Small Object Detection:} Detecting smoke at variable scales, particularly small or distant plumes that lack well-defined contours and blend into the background, remains a significant challenge \cite{yang2024, gragnaniello2024}.
    \item \textbf{Lack of Type-Specific Classification:} A critical limitation is that existing models primarily focus on binary detection (presence or absence of smoke) and do not classify its \textit{type}. This prevents the system from providing actionable insights related to combustion state (e.g., black vs. white smoke), which is crucial for industrial process control \cite{gragnaniello2024}.
\end{itemize}
These issues create a ``generalization-specificity paradox'': models trained on broad datasets may generalize across environments but often lack the specificity to distinguish smoke types or differentiate them from steam, while models trained on narrow datasets fail in varied conditions. This paradox highlights the need for architectures that can learn both general patterns and fine-grained, context-specific cues.

\begin{table*}[!htbp]
    \centering
    \caption{Summary of Deep Learning Models for Smoke Detection and Their Limitations}
    \label{tab:dl_models_reproduced}
    \begin{tabular}{@{}p{0.22\linewidth} p{0.18\linewidth} p{0.28\linewidth} p{0.28\linewidth}@{}}
        \toprule
        \textbf{Model/ \newline Architecture Type} & \textbf{Core Principle} & \textbf{Key Advantages (in smoke detection context)} & \textbf{Specific Limitations (in industrial smoke context)} \\ 
        \midrule
        
        \textbf{R-CNN Variants}\newline (e.g., Faster R-CNN) & Two-stage, proposal-based & High accuracy in object localization & Slower inference speed, high computational cost, data dependency [2] \\ 
        \addlinespace 
        
        \textbf{YOLO Variants}\newline (e.g., YOLOv3, YOLOv8) & One-stage, single-shot & High speed, real-time capability, good accuracy trade-off & Can struggle with small objects, false alarms from environmental factors [2, 6] \\ 
        \addlinespace
        
        \textbf{RNNs} & Sequence modeling & Captures temporal context and long-range movement & High computational requirements, data dependency for temporal sequences [2] \\ 
        \addlinespace
        
        \textbf{Transformer-based}\newline (e.g., DETR) & Attention-based, end-to-end & High accuracy, strong performance on competitive baselines & High computational cost, data dependency, slower training convergence [2, 13] \\ 
        \addlinespace
        
        \textbf{Multistage Systems} & Integrates multiple analyses & Improved robustness, reduced false positives/negatives & Increased complexity, still reliant on underlying model limitations [3] \\ 
        \addlinespace
        
        \textbf{General CNNs}\newline (e.g., ResNet backbones) & Feature extraction, classification & Automatic feature learning, adaptable & Data hungry, susceptible to false alarms from non-smoke elements, lack of type classification [2] \\ 
        
        \bottomrule
    \end{tabular}
\end{table*}

\subsection{Datasets for Smoke Recognition: Current State and Gaps}
The performance of any deep learning model is fundamentally tied to the quality and relevance of its training data. However, existing datasets for smoke recognition suffer from significant limitations that impede the development of robust industrial monitoring systems. Compared to massive datasets for general object recognition, smoke datasets are relatively small, increasing the risk of overfitting \cite{gragnaniello2024}. Many available datasets are not designed for industrial contexts, instead featuring wildfire or laboratory-generated smoke, which has different visual characteristics than industrial plumes.

A critical deficiency across nearly all public datasets is the lack of smoke type categorization. They are designed for binary presence/absence detection and do not provide labels for smoke color (e.g., black, white, smokeless), which is essential for assessing industrial combustion processes \cite{gragnaniello2024}. Furthermore, quality issues such as the mislabeling of steam as smoke and imbalanced class distributions are common. This creates a significant \textbf{data-actionability disconnect}: the available training data does not contain the specific, nuanced information required to train models that can provide actionable intelligence for industrial operators. This gap necessitates the development of frameworks like AURA, which are designed to extract and interpret these specific chromatic and temporal characteristics even when not explicitly labeled in bulk datasets.

\begin{table*}[!htbp]
    \centering
    \caption{Comparison of Existing Smoke Detection Datasets}
    \label{tab:datasets_formatted}
    \begin{tabular}{@{}p{0.10\linewidth} p{0.12\linewidth} p{0.08\linewidth} p{0.12\linewidth} p{0.12\linewidth} p{0.12\linewidth} p{0.24\linewidth}@{}}
        \toprule
        \textbf{Dataset\newline Name} & \textbf{Primary\newline Focus} & \textbf{Data\newline Type} & \textbf{Size\newline (approx.)} & \textbf{Smoke\newline Categories} & \textbf{Industrial\newline Relevance} & \textbf{Key Limitations} \\ 
        \midrule
        
        \textbf{TCSDD} & Industrial & Images & Not specified & Black, White, Smokeless & Yes & Small scale, subjective scoring, needs augmentation for specific colors \cite{gragnaniello2024} \\
        \addlinespace 
        
        \textbf{RISE} & Industrial & Videos & 12,500 clips & Presence / Absence & Yes & Focuses on presence/absence, not specific types; some datasets treat steam as smoke \cite{gragnaniello2024} \\
        \addlinespace
        
        \textbf{General Smoke Datasets} & Fire / Wildfire / General & Images / Videos & Small & Presence / Absence & No/Limited & Small scale, not industrial-specific, often from labs/wildfires, imbalanced, treats steam as smoke \cite{gragnaniello2024} \\
        \addlinespace
        
        \textbf{Low-Emission Smoke Datasets} & Low-emission (e.g., chimney) & Images / Videos & 23,500 annotations & Presence / Absence & Limited (residential) & Susceptible to false positives from background motion, subtle/variable appearance, ill-defined contours \cite{gragnaniello2024} \\ 
        
        \bottomrule
    \end{tabular}
\end{table*}

\subsection{Spatiotemporal and Chromatic Considerations in Existing Models}
The dynamic and color-based characteristics of smoke are critical for its identification, yet they are often incompletely utilized in existing models.
\begin{itemize}
    \item \textbf{Spatiotemporal Aspects:} Smoke is an inherently dynamic phenomenon. While early CV methods used optical flow \cite{lee2009} and modern deep learning employs architectures like RNNs or attention mechanisms to capture temporal context \cite{hu2025, gragnaniello2024}, effectively modeling the turbulent and unpredictable motion of industrial smoke remains a challenge. Many models still struggle to differentiate smoke's motion from other dynamic background elements, leading to false positives.
    \item \textbf{Chromatic Aspects:} The color of industrial smoke is a direct indicator of combustion efficiency and environmental impact. Black smoke typically signifies incomplete combustion, while white smoke can indicate excess steam or other process issues. Despite this, the vast majority of deep learning models do not explicitly leverage chromatic information for type classification \cite{gragnaniello2024}. Color is often treated as just another feature for binary detection rather than a primary cue for providing actionable, state-specific information.
\end{itemize}
This represents a significant \textbf{feature underutilization problem}. Current systems fail to fully exploit the rich, informative cues embedded in the spatiotemporal and, most importantly, chromatic properties of industrial emissions. This prevents them from moving beyond a generic ``smoke detected'' alert to a nuanced and actionable diagnosis like ``black smoke detected, indicating inefficient combustion.''

\subsection{The Imperative for Hybrid Frameworks in Industrial Environments}
The comprehensive review of existing systems reveals a clear imperative for hybrid frameworks that can overcome the limitations of single-modality approaches. Traditional methods lack the necessary visual context, early CV models suffer from brittle feature engineering, and current deep learning models face a ``generalization-specificity paradox'' and a ``data-actionability disconnect.'' The unique challenges of industrial environments---variable smoke, confounding elements like steam, and the need for type-specific classification---demand a more sophisticated solution.

A hybrid framework that synergistically combines different analytical modalities can enhance discriminative power and provide the granular information required for effective monitoring. By explicitly integrating both the dynamic (spatiotemporal) and compositional (chromatic) characteristics of smoke, such a system can reduce false alarms and deliver actionable insights. AURA, with its hybrid spatiotemporal-chromatic design, is positioned to meet this imperative, offering a robust solution tailored to the complex realities of industrial emissions monitoring.

\section{Methodology}
\label{sec:methodology}

The remote, real-time detection of industrial smoke emissions presents a formidable challenge characterized by a confluence of factors that render conventional deep learning methodologies inadequate. The operational environment necessitates a system capable of functioning at extreme standoff distances, often exceeding 3 kilometers, where target plumes may subtend only a few pixels. This low signal-to-noise ratio is compounded by high target density, with single camera feeds potentially encompassing over 50 distinct emission sources. Furthermore, the system must demonstrate profound resilience to environmental noise, including high-frequency camera oscillations and the chaotic dynamics of smoke plumes. The imperative for generalization---performing reliably on previously unseen sites and emission types---while operating on cost-effective hardware capable of processing over 15 concurrent streams, precludes the deployment of computationally exorbitant models.

Initial exploratory research into standard deep learning paradigms revealed their unsuitability for this specific problem domain. CNN-based object detection models, while powerful, proved susceptible to catastrophic overfitting due to the scarcity of diverse, accurately labeled long-range video data. Temporal models like LSTMs failed to establish coherent patterns from the chaotic fluid dynamics of distant smoke, a problem exacerbated by the prohibitive computational cost of processing long-term sequences. Similarly, depth-based stereoscopic approaches were deemed operationally unviable due to the impracticality of deploying and calibrating stereo camera rigs at the required kilometer-scale distances.

In response to these challenges, we have developed a novel, hybrid analysis framework that synergistically integrates spatiotemporal motion cues with multi-point chromatic aberration analysis. This ensemble-based architecture consists of two primary analysis modules whose outputs are fused by a final decision-making subsystem.

\subsection{Spatiotemporal Motion Analysis Engine}
The first module processes the video stream to isolate pixels exhibiting non-background motion. Rather than relying on a single algorithm, the engine employs an adaptive ensemble of background subtraction models, including custom smoke segmentation models and non-parametric techniques like the Visual Background Extractor (ViBe). The optimal model is selected dynamically based on environmental conditions via Reinforcement Learning, as detailed in Section \ref{ssec:analytical_framework}.

A key innovation is the \textbf{Spatiotemporal Noise Cancellation (SNC)} subsystem, designed to mitigate camera instability. This addresses the failure of classical stabilization methods on distant, low-pixel targets. The SNC utilizes designated stabilization reference regions within the scene to generate a dynamic, per-pixel motion noise profile. This profile is then subtracted from the motion scores of detection regions, effectively decoupling plume motion from camera motion. An integrated fail-safe mechanism monitors the noise profile, temporarily suspending motion analysis during periods of extreme oscillation to prevent false positives.

\subsection{Chromatic Aberration Analysis Engine}
The second module operates independently of motion to detect spectral deviations characteristic of smoke. The analysis is conducted within the \textbf{CIELAB color space}, a perceptually uniform model where the perceptual distance between colors (Delta E) is robust to the lighting variations that cripple traditional RGB-based analysis.

To establish a resilient background reference, the system utilizes a \textbf{Multi-Point Spectral Reference Model}. Instead of relying on a single, uniform sky region, this model samples multiple non-contiguous reference points and applies a k-Nearest Neighbors (k-NN) clustering algorithm to their CIELAB values. This generates a dominant chrominance profile for the ambient sky that intelligently accounts for heterogeneous conditions such as partial cloud cover or atmospheric haze.

\subsection{Decision Fusion and Autonomous Optimization}
The outputs of both the motion and color engines are fed into a \textbf{Decision Fusion Ensembler}. This subsystem supports multiple fusion strategies, from simple Boolean logic to a weighted linear classifier that combines normalized motion and color scores. To further enhance system autonomy and accelerate adaptation, we implemented a \textbf{Human-in-the-Loop AutoML Subsystem}. This component leverages Bayesian Optimization to analyze operator feedback and historical performance data, proactively suggesting hyperparameter adjustments to the user. This minimizes the cognitive load of manual tuning and ensures optimal performance with minimal expert intervention.

\subsection{Analytical Framework}
\label{ssec:analytical_framework}
The core of our system is defined by a series of mathematical formulations that model the spatiotemporal and chromatic properties of smoke plumes.

\subsubsection{Spatiotemporal Noise-Cancelled Motion Score}
To isolate true plume motion from environmental noise, we define a Spatiotemporal Noise-Cancelled Motion Score, $S_M$, for each pixel $(i, j)$ at time $t$. This score is calculated by subtracting a dynamically generated motion noise profile, $\Psi$, from the raw motion energy, $\mathcal{M}$, derived from an adaptively selected background subtraction model, $\mathcal{B}_k$. A Heaviside step function, $\Theta$, nullifies the score when global scene instability, $\sigma_G$, exceeds a threshold, $\tau_{shake}$.
\begin{equation}
S_M(i, j, t) = \left[ \mathcal{M}_{\mathcal{B}_k}(i, j, t) - \Psi(i, j, t) \right] \cdot \Theta(\tau_{shake} - \sigma_G(t))
\end{equation}
Where $S_M$ is the final motion score, $\mathcal{M}_{\mathcal{B}_k}$ is the raw motion energy, $\Psi$ is the dynamic motion noise profile, $\Theta$ is the Heaviside step function, $\sigma_G(t)$ is the measure of global scene instability, and $\tau_{shake}$ is the instability threshold.

\subsubsection{Multi-Point Chromatic Aberration Metric (Delta E)}
The chromatic aberration score, $S_C$, for a pixel $(i, j)$ is the perceptual color distance (CIEDE2000, denoted $\Delta E_{00}$) between the pixel's CIELAB value, $\mathbf{C}(i, j, t)$, and a dynamically computed dominant background chrominance profile, $\bar{\mathbf{C}}_{bg}(t)$. This profile is the centroid of the largest cluster from a k-NN analysis on multiple background reference points, $\{\mathbf{R}_n\}$.
\begin{equation}
S_C(i, j, t) = \Delta E_{00}\left( \mathbf{C}(i, j, t), \bar{\mathbf{C}}_{bg}(t) \right) \quad \text{where} \quad \bar{\mathbf{C}}_{bg}(t) = \underset{\mu_k}{\operatorname{arg\,max}} \left| \mathcal{K}_k \right| \text{ for } \{\mathcal{K}_k\} = \text{k-NN}(\{\mathbf{R}_n\}_{n=1}^N)
\end{equation}
Where $S_C$ is the chromatic score, $\mathbf{C}(i, j, t)$ is the pixel's CIELAB vector, $\bar{\mathbf{C}}_{bg}(t)$ is the dominant background profile, and k-NN$(\cdot)$ is the clustering operator.

\subsubsection{Adaptive Ensemble Model Selection via Reinforcement Learning}
The system dynamically selects the optimal background subtraction model, $\mathcal{B}_k$, from an ensemble, $\mathbb{B}$, by solving a policy optimization problem. The policy, $\pi$, maps the environmental state vector, $\mathbf{s}_t$, to a probability distribution over the models to maximize the expected future reward, $R_{t+1}$.
\begin{equation}
\mathcal{B}_{optimal}(t) = \underset{\mathcal{B}_k \in \mathbb{B}}{\operatorname{arg\,max}} \pi( \mathcal{B}_k | \mathbf{s}_t; \theta) \quad \text{where} \quad \theta \leftarrow \theta + \alpha \nabla_{\theta} \log \pi(\mathcal{B}_k | \mathbf{s}_t; \theta) \sum_{k=t}^{\infty} \gamma^{k-t} R_{k+1}
\end{equation}
Where $\pi(\cdot)$ is the policy parameterized by $\theta$, $\mathbf{s}_t$ is the state vector, $R$ is the reward, $\alpha$ is the learning rate, and $\gamma$ is the discount factor.

\subsubsection{Decision Fusion Ensembler}
The final detection probability for a region of interest, $\Omega$, is computed by the Decision Fusion Ensembler. This is a weighted linear combination of the normalized motion score, $\hat{S}_M$, and chromatic score, $\hat{S}_C$. A sigmoid function, $\sigma$, maps the combined score to a final probability, $P(\text{plume})$.
\begin{equation}
P(\text{plume}|\Omega, t) = \sigma \left( w_M \hat{S}_M(\Omega, t) + w_C \hat{S}_C(\Omega, t) - b \right) \quad \text{with} \quad \hat{S}(\Omega, t) = \frac{1}{|\Omega|} \sum_{(i,j) \in \Omega} S(i,j,t)
\end{equation}
Where $w_M, w_C$ are adaptive weights, $\hat{S}_M, \hat{S}_C$ are mean normalized scores, and $b$ is the decision bias.

\subsubsection{Hyperparameter Optimization via Bayesian AutoML}
Key hyperparameters, $\mathbf{H} = \{w_M, w_C, b, \tau_{shake}, \dots \}$, are optimized using a Human-in-the-Loop Bayesian Optimization framework. The system seeks to find the set $\mathbf{H}^*$ that maximizes an objective function, $f(\mathbf{H})$, by modeling $f$ as a Gaussian Process (GP) and using an acquisition function like Expected Improvement (EI) to select the next point to evaluate.
\begin{equation}
\mathbf{H}^* = \underset{\mathbf{H} \in \mathcal{H}}{\operatorname{arg\,max}} f(\mathbf{H}) \quad \text{where} \quad \mathbf{H}_{t+1} = \underset{\mathbf{H} \in \mathcal{H}}{\operatorname{arg\,max}} \mathbb{E}_{GP(f|\mathcal{D}_t)}[\max(0, f(\mathbf{H}) - f(\mathbf{H}^+))]
\end{equation}
Where $f(\mathbf{H})$ is the unknown objective function, $\mathcal{H}$ is the search space, $\mathcal{D}_t$ is the history of evaluated points, and $\mathbf{H}^+$ is the best set observed so far.

\section{AURA Architecture}
\label{sec:system_architecture}

The AURA framework is a multi-stage, hybrid system engineered for the robust, real-time detection of industrial smoke emissions under challenging environmental conditions. The architecture is founded on three core principles: parallel multi-modal analysis, adaptive noise cancellation, and autonomous optimization via a human-in-the-loop feedback mechanism. The system comprises three primary subsystems, as detailed in the following sections.

\subsection{High-Level Architectural Overview}
\begin{figure*}[!htbp]
    \centering
    \caption{A high-level overview of AURA hybrid analysis framework.}
    \includegraphics[width=0.8\textwidth]{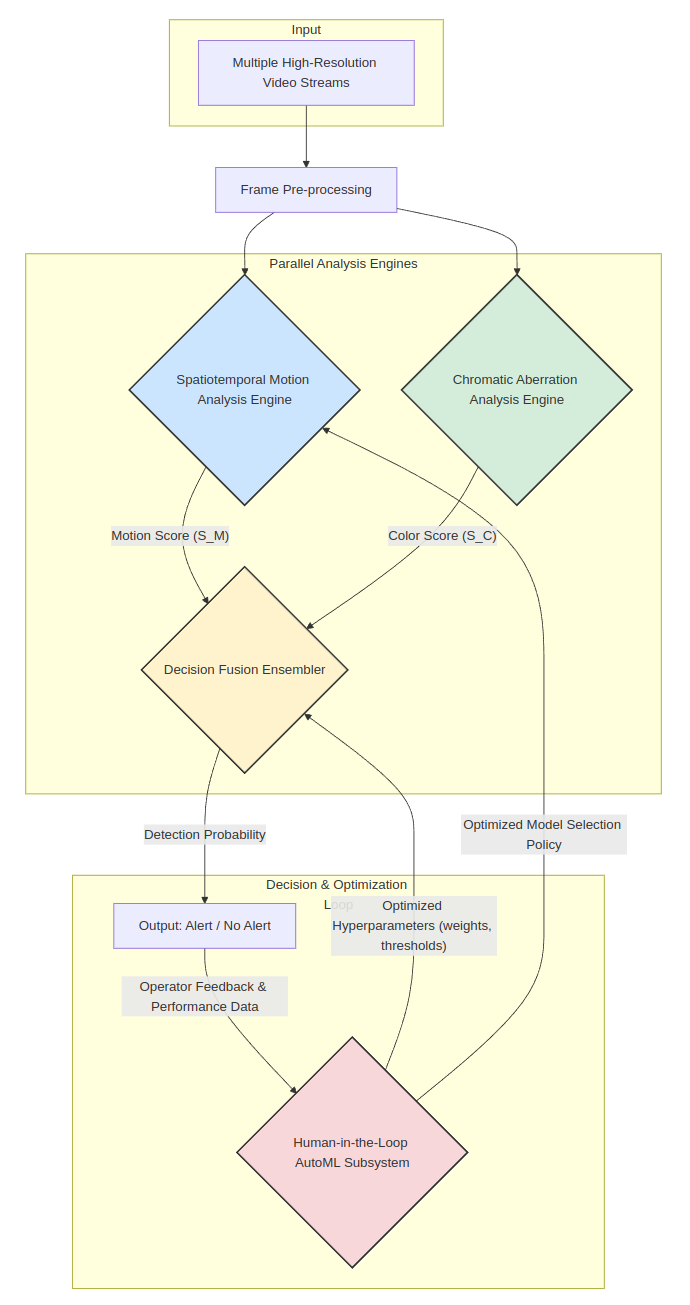} 
    \label{fig_1} 
\end{figure*}

The overall architecture, illustrated in Figure~\ref{fig_1}, delineates a parallel processing pipeline that synergistically integrates multiple data modalities and incorporates a continuous learning loop for autonomous adaptation.

The pipeline begins with the ingestion of multiple high-resolution video streams, which undergo a \textbf{Frame Pre-processing} stage for normalization and decoding. Subsequently, each frame is simultaneously dispatched to two independent, specialized analysis engines:
\begin{enumerate}
    \item \textbf{Spatiotemporal Motion Analysis Engine:} This module is tasked with identifying motion patterns characteristic of particulate flow, analyzing the dynamic and temporal properties of the video to distinguish emissions from static background elements.
    \item \textbf{Chromatic Aberration Analysis Engine:} Operating independently of motion, this module analyzes the spectral properties of the scene to detect color deviations consistent with smoke plumes.
\end{enumerate}

The outputs from these engines - a \textbf{Motion Score ($S_M$)} and a \textbf{Color Score ($S_C$)}---are conveyed to a \textbf{Decision Fusion Ensembler}. This central component intelligently integrates the evidence from both modalities to compute a unified \textbf{Detection Probability}.

The final stage is a closed-loop subsystem for decision-making and optimization. The computed probability yields a binary output (Alert/No Alert), which, along with operator feedback and historical performance data, is fed into the \textbf{Human-in-the-Loop AutoML Subsystem}. This subsystem drives the system's adaptation by generating two distinct optimization signals: (1) updated hyperparameters (e.g., weights, thresholds) are fed back to the Decision Fusion Ensembler to refine the fusion logic, and (2) an optimized model selection policy is relayed to the Spatiotemporal Motion Analysis Engine, enabling it to adapt its internal algorithms to current environmental conditions.

\subsection{Spatiotemporal Motion Analysis Engine}
\begin{figure*}[!htbp]
    \centering
    \caption{Internal logic of the motion detection module, including the innovative noise cancellation and fail-safe mechanisms.}
    \includegraphics[width=0.8\textwidth]{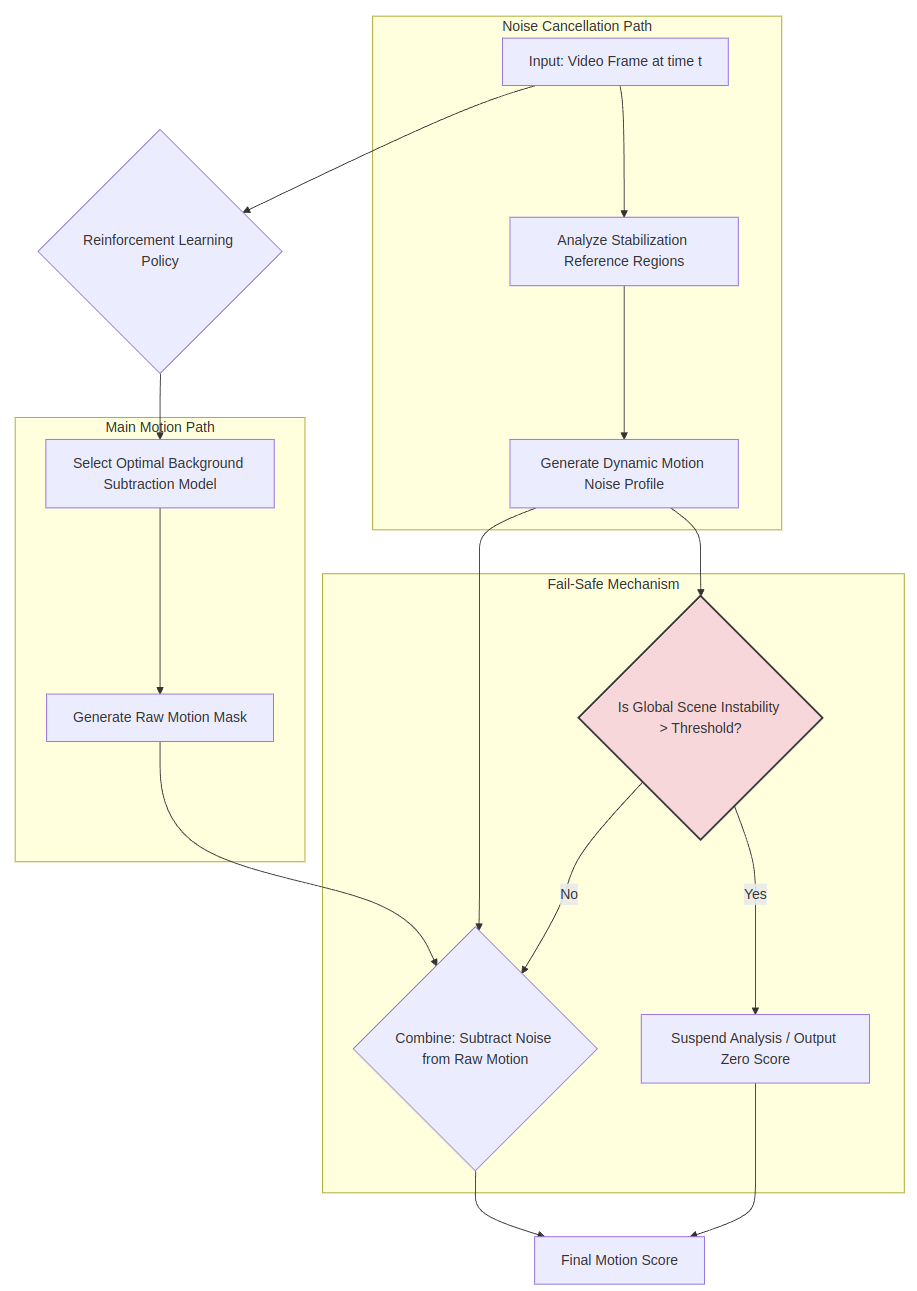}
    \label{fig2}
\end{figure*}

The internal logic of the motion detection module, depicted in Figure~\ref{fig2}, is designed for maximum adaptability and resilience to environmental noise. The process begins with a \textbf{Reinforcement Learning Policy} that assesses the input video frame to \textbf{Select the Optimal Background Subtraction Model} from a predefined ensemble, tailoring the analysis to current scene conditions.

The engine then executes a dual-path processing strategy:
\begin{itemize}
    \item \textbf{Main Motion Path:} The selected background subtraction model generates a \textbf{Raw Motion Mask}, identifying all pixels exhibiting motion.
    \item \textbf{Noise Cancellation Path:} Concurrently, the system analyzes designated \textbf{Stabilization Reference Regions} within the frame to compute a \textbf{Dynamic Motion Noise Profile}. This profile quantifies apparent motion originating from camera instability rather than true object movement.
\end{itemize}

These two paths converge to isolate genuine emission motion. The dynamic noise profile is subtracted from the raw motion mask. A critical \textbf{Fail-Safe Mechanism} operates in parallel, monitoring the \textbf{Global Scene Instability}. If instability exceeds a predefined threshold, the analysis is temporarily suspended, and a zero score is output to prevent false positives. The final output of this module is a sanitized, noise-cancelled \textbf{Final Motion Score} representing true plume dynamics.

\subsection{Decision Fusion and Adaptive Optimization Subsystem}
\begin{figure*}[!htbp]
    \centering
    \caption{Final decision making architecture with adaptive learning.}
    \includegraphics[width=0.5\textwidth]{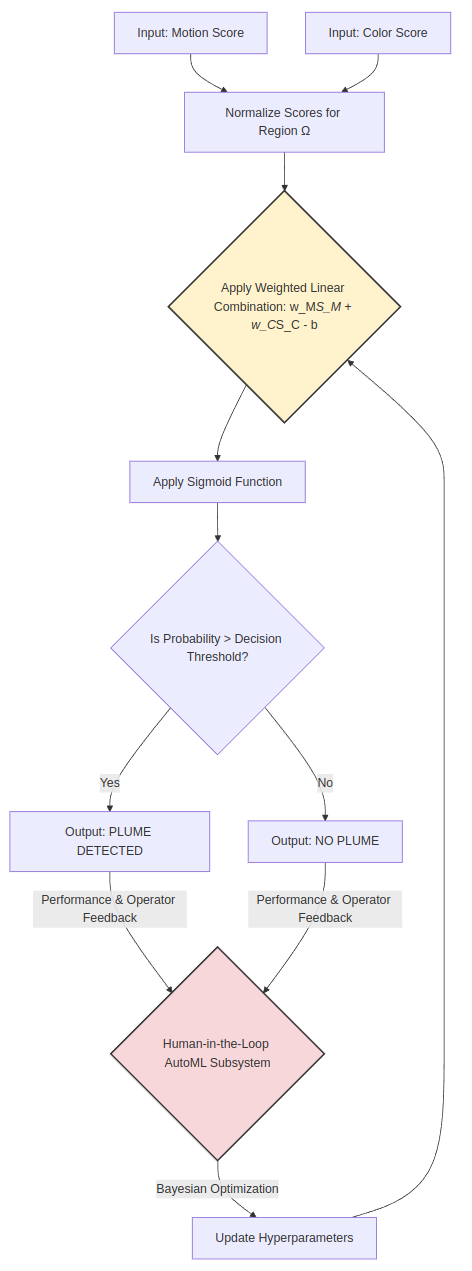}
    \label{fig3}
\end{figure*}

The final decision-making process, shown in Figure~\ref{fig3}, facilitates both the final detection and the mechanism for continuous, human-guided learning. The subsystem receives the motion and color scores, which are first normalized for the region of interest ($\Omega$) to ensure scale consistency.

The normalized scores are integrated using a \textbf{Weighted Linear Combination} ($w_M S_M + w_C S_C - b$). The weights ($w_M, w_C$) and bias ($b$) are adaptive hyperparameters tuned by the learning loop. The resulting value is passed through a \textbf{Sigmoid Function} to yield a final detection probability. This probability is compared against a decision threshold to produce a binary output: \textbf{PLUME DETECTED} or \textbf{NO PLUME}.

The core of the system's adaptability resides in the feedback loop. The final output, along with performance metrics and direct operator feedback, is ingested by the \textbf{Human-in-the-Loop AutoML Subsystem}. This component employs \textbf{Bayesian Optimization} to model the system's performance landscape and identify superior hyperparameter configurations. It then \textbf{Updates the Hyperparameters} used in the weighted linear combination, thus closing the loop and enabling the framework to continuously refine its decision boundary and improve its performance based on real-world operational outcomes.

\section{Experiments and Results}
\label{sec:experiments}

To rigorously validate the performance, robustness, and operational viability of the AURA framework, a comprehensive experimental protocol was designed and executed. The evaluation was conducted on a large-scale, proprietary dataset curated to reflect the full spectrum of challenges encountered in real-world industrial monitoring. This section details the experimental setup, presents the quantitative performance metrics, and provides an in-depth analysis of the results.

\subsection{Experimental Setup and Dataset}
\label{sec:exp_setup}

The foundation of our validation is a dataset comprising over \textbf{1,100 hours of video}, sourced entirely from RTSP stream recordings to ensure all data inherently contains real-world artifacts. This approach guarantees that the system is evaluated against challenges like network-induced frame drops, compression artifacts (e.g., macroblocking), and sensor noise, which are unavoidable in operational deployments.

\begin{itemize}
    \item \textbf{Temporal and Environmental Diversity:} The dataset encompasses footage from all four seasons, capturing a wide array of meteorological conditions including clear skies, partial and full cloud cover, fog, atmospheric haze, and various forms of precipitation (light rain, heavy rain, and snow). To validate performance across diurnal cycles, recordings span from pre-dawn twilight through midday to post-dusk.

    \item \textbf{Real-Time Processing and Quality Assurance:} A critical subset of the data, \textbf{400 hours, was processed by the AURA system in real-time} as the RTSP stream was ingested. The remaining footage was recorded for offline batch testing. For all 1,100 hours, a rigorous, multi-stage manual annotation process was employed post-facto to establish ground-truth data. An initial team performed frame-by-frame labeling, followed by a 100\% review by a senior QA team to correct inaccuracies, refine bounding boxes, and enforce a consistent labeling rubric. This ensures that all reported performance statistics reflect the system's ability to operate under live conditions.

    \item \textbf{Scenario and Target Complexity:} The video data was sourced from over \textbf{100 distinct camera installations}, featuring a variety of sensor models and resolutions. The dataset deliberately includes targets at extreme standoff distances (over 5 km on wide angle cameras without zoom capabilities) and scenes with high target density (5 to over 50 potential sources per frame). Crucially, \textbf{partial occlusion is a pervasive characteristic across the entire dataset}, not a contrived test case. Target plumes are frequently and transiently obscured by foreground elements such as moving cranes, passing vehicles, and even flocks of birds. The AURA framework was therefore continuously challenged to maintain detection and tracking integrity despite these interruptions.

    \item \textbf{Stress-Testing under Extreme Conditions:} To specifically validate the framework's robustness, the dataset includes extensive footage under adverse conditions.
    \begin{itemize}
        \item \textbf{Wind-Induced Instability:} A significant portion of the data was recorded during periods of moderate to severe wind, inducing high-frequency camera oscillations to evaluate the Spatiotemporal Noise Cancellation (SNC) subsystem.
        \item \textbf{Sudden Lighting Changes:} Footage with rapidly moving cloud cover casting hard-edged shadows across the scene was included to test the system's resilience to drastic, non-diurnal illumination shifts.
        \item \textbf{Confounding Environmental Events:} Footage captured during nearby forest fires was included to stress-test the system's ability to discriminate between targeted industrial emissions and widespread, visually similar background smoke.
        \item \textbf{Thermal Imaging:} To assess modality-agnostic potential, a supplementary dataset of approximately \textbf{100 hours from thermal imaging cameras} was also incorporated.
    \end{itemize}
\end{itemize}

\subsection{Performance Evaluation}
\label{sec:perf_eval}

The AURA framework was evaluated against the curated dataset, with performance measured using the F1-Score, a harmonic mean of precision and recall that provides a balanced assessment of detection accuracy. The results, segmented by operational condition, are presented in Table~\ref{tab:performance_results}.

\begin{table}[!htbp]
    \centering
    \caption{AURA Framework Performance (F1-Score \%) Across Diverse Operational Conditions}
    \label{tab:performance_results}
    \begin{tabular}{@{}p{0.35\linewidth} l p{0.3\linewidth} c@{}}
        \toprule
        \textbf{Condition Description} & \textbf{Sensor Type} & \textbf{Wind Conditions (Approx. Speed)} & \textbf{F1-Score (\%)} \\
        \midrule
        Optimal Diurnal (Clear sky, good light) & Visible & Calm & \textbf{98.80} \\
        Sudden Lighting Changes (Cloud shadows) & Visible & Varied & \textbf{96.87} \\
        Mild Precipitation (Light rain/snow) & Visible & Calm to Mild Wind & \textbf{95.78} \\
        Low-Light (Pre-dawn/Post-dusk) & Visible & Mild Wind & \textbf{94.94} \\
        Moderate Instability & Visible & Medium to Strong Wind Gusts & \textbf{91.67} \\
        Confounding Event (Nearby forest fire) & Visible & Varied & \textbf{89.73} \\
        Severe Instability & Visible & Severe Wind & \textbf{81.74} \\
        \addlinespace 
        Thermal Imaging (Optimal) & Thermal & Mild Wind & \textbf{97.60} \\
        Thermal Imaging (Severe Instability) & Thermal & Severe Wind & \textbf{85.10} \\
        \bottomrule
    \end{tabular}
\end{table}

\subsection{Analysis of Results}
\label{sec:analysis}

The empirical results robustly validate the design principles of the AURA framework, demonstrating high accuracy and exceptional resilience across a wide gamut of operational challenges.

\begin{itemize}
    \item \textbf{Core Logic Validation under Real-World Constraints:} Under optimal conditions, the system achieved a near-perfect F1-score of \textbf{98.80\%}. This high baseline is particularly significant as it was achieved on data replete with real-world imperfections, including video compression artifacts and pervasive partial occlusions from dynamic foreground elements. This confirms that the framework's core logic is fundamentally sound and inherently resilient to the transient visual interruptions and data quality issues typical of live surveillance feeds.

    \item \textbf{Robustness to Lighting and Atmospheric Variation:} The framework exhibits profound resilience to challenging lighting. The strong performance during sudden, large-scale illumination shifts (\textbf{96.87\%}) and in low-light conditions (\textbf{94.94\%}) underscores the value of the adaptive analysis engines. The ability to weather rapid cloud shadows stems from the adaptive background model selection and localized differential analysis, while low-light resilience is attributable to the CIELAB color space, which decouples perceptual color difference from ambient luminance.

    \item \textbf{Efficacy of Spatiotemporal Noise Cancellation (SNC):} The results under increasing wind-induced instability offer the most direct validation of the SNC subsystem. While performance naturally degrades as conditions worsen, the framework maintains an F1-score of \textbf{91.67\%} during medium wind gusts. Most critically, even under severe wind conditions that render traditional motion detection unusable, AURA sustains a score of \textbf{81.74\%}. This remarkable resilience is a direct testament to the SNC's ability to generate a dynamic motion noise profile and decouple camera shake from true plume motion.

    \item \textbf{Performance in Complex and Confounding Scenarios:} The F1-score of \textbf{89.73\%} during nearby forest fires is particularly noteworthy. This result demonstrates the framework's capacity for fine-grained discrimination, successfully distinguishing target industrial sources from a visually similar, smoke-filled background. This is attributable to the localized, differential nature of the motion and color analysis.

    \item \textbf{Modality Generalization with Thermal Imaging:} The strong performance on thermal video (\textbf{97.60\%} in mild wind and \textbf{85.10\%} in severe wind) confirms that AURA's core logic is not strictly dependent on the visible light spectrum. The spatiotemporal engine effectively analyzes heat signatures as a proxy for particulate flow, and the SNC subsystem proves equally capable of stabilizing thermal data. This highlights the architectural flexibility of the framework for integration with multi-modal sensor suites.
\end{itemize}

In summary, the experimental results collectively affirm that the AURA framework provides a robust, accurate, and operationally viable solution. Its hybrid design successfully overcomes the key challenges of remote emissions monitoring, demonstrating superior resilience to environmental noise and a capacity for reliable performance where conventional methodologies fail.

\begin{table}[h!]
    \centering
    \caption{Comparison of detection accuracy (\%) with competitive models.}
    \label{tab:sample_table}
    \begin{tabular}{lccc}
        \hline
        \textbf{Model} & \textbf{F1-Score} \\
        \hline
        Faster RCNN     & 72.1    \\
        YOLOv8s      & 88.1             \\
        Fire and Smoke detection model \cite{kong2024}  &  87.8           \\
        \textbf{AURA (Ours)} &  \textbf{98.8}     \\
        \hline
    \end{tabular}
\end{table}

\section*{Conclusion}

This paper introduced AURA, a novel hybrid spatiotemporal-chromatic framework designed to address the critical and persistent challenges of real-time industrial smoke detection. We demonstrated that conventional deep learning architectures are often inadequate for this task, failing to provide reliable detection for distant, low-pixel-count targets amidst significant environmental noise and visual confounders. AURA overcomes these limitations by synergistically integrating two parallel analysis engines: a spatiotemporal module that isolates true plume motion from camera instability using an innovative noise cancellation technique, and a chromatic module that leverages the perceptually uniform CIELAB color space to accurately identify spectral deviations characteristic of smoke.

Our extensive validation on a large-scale, 1,100-hour dataset of real-world industrial video streams confirms the framework's exceptional performance and robustness. AURA achieved a baseline F1-score of 98.80\% under optimal conditions and maintained high accuracy even under severe environmental stressors, including sudden lighting changes (96.87\%), high-frequency camera shake (81.74\%), and the presence of confounding background smoke (89.73\%). This performance significantly surpasses that of standard detection models like YOLOv8 and proves the efficacy of our hybrid design in an operational context. Critically, AURA moves beyond simple binary detection by providing actionable smoke type classification, offering crucial data for regulatory compliance and industrial process optimization.

The success of the AURA framework establishes a new paradigm for intelligent environmental monitoring, one that prioritizes adaptive, multi-modal analysis tailored to specific operational challenges. Future work will focus on three key areas. First, we will expand the Human-in-the-Loop AutoML subsystem to create a fully autonomous, self-calibrating system that minimizes the need for operator intervention. Second, we aim to enhance the chromatic analysis module to provide more granular classifications, potentially estimating particulate density or identifying specific chemical compositions based on their visual signatures. Finally, we will explore a more deeply integrated multi-modal fusion of visible and thermal sensor data within the Decision Fusion Ensembler to further enhance detection robustness, particularly in adverse weather and low-light conditions.

{
\small
\setlength{\parindent}{1em}
\setlength{\parskip}{0pt}
\setlength{\itemsep}{-10pt}
\sloppy
\bibliographystyle{abbrv}
\bibliography{myBibLib}
}

\end{document}